\documentclass[11pt, letterpaper]{article}

\usepackage{amsmath}
\usepackage{amssymb}
\usepackage{amsthm}
\usepackage{graphicx}
\usepackage{hyperref}
\usepackage{url}
\usepackage{algorithm}
\usepackage{algorithmic}
\usepackage{caption}
\usepackage{subcaption}
\usepackage{booktabs}
\usepackage{authblk}
\usepackage{appendix}
\usepackage[utf8]{inputenc}
\usepackage[T1]{fontenc}
\usepackage[english]{babel}
\usepackage{geometry}
\usepackage{xcolor}
\usepackage{fancyhdr}
\usepackage{microtype}
\usepackage{parskip}
\usepackage{tabularx}
\usepackage{enumitem}

\usepackage{titlesec}
\usepackage{indentfirst}

\titleformat{\subsubsection}[hang]{\bfseries\normalsize}{\hspace{0em}}{0pt}{}

\usepackage{float}
\usepackage[numbers,sort]{natbib}
\hypersetup{
    colorlinks=true,
    linkcolor=blue,
    filecolor=magenta,      
    urlcolor=cyan,
    citecolor=red,
}

\newcolumntype{C}{>{\centering\arraybackslash}X} 
\usepackage[normalem]{ulem}

\hypersetup{
    colorlinks=true,
    linkcolor=blue,
    filecolor=magenta,      
    urlcolor=cyan,
    citecolor=red,
}

\definecolor{headercolor}{RGB}{0, 50, 100}
\title{GP-GPT: Large Language Model for Gene-Phenotype Mapping}

\author[1]{Yanjun Lyu}
\author[2]{Zihao Wu}
\author[3]{Lu Zhang}
\author[1]{Jing Zhang}
\author[2]{Yiwei Li}
\author[2]{Wei Ruan}
\author[2]{Zhengliang Liu}
\author[1]{Zeyu Zhang}
\author[4]{Xiang Li}
\author[5]{Rongjie Liu}
\author[6]{Chao Huang} 
\author[7]{Wentao Li}
\author[2]{Tianming Liu}
\author[1]{Dajiang Zhu}

\affil[1]{Department of Computer Science and Engineering, The University of Texas at Arlington, Arlington, TX 76015, USA}
\affil[2]{School of Computing, University of Georgia, Athens, GA 30602, USA}
\affil[3]{Department of Computer Science, Indiana University Indianapolis, IN 46202, USA}
\affil[4]{Department of Radiology, Massachusetts General Hospital and Harvard Medical School, Boston, MA 02115, USA}
\affil[5]{Department of Statistics, University of Georgia, Athens, GA 30602, USA}
\affil[6]{Department of Epidemiology \& Biostatistics, University of Georgia, GA 30602, USA}
\affil[7]{Department of Environmental Health Science, University of Georgia, GA 30602, USA}

\date{}

\begin{document}

\maketitle

\begin{abstract}

Pre-trained large language models(LLMs) have attracted increasing attention in biomedical domains due to their success in natural language processing. However, the complex traits and heterogeneity of multi-sources genomics data pose significant challenges when adapting these models to the bioinformatics and biomedical field. To address these challenges, we present GP-GPT, the first specialized large language model for genetic-phenotype knowledge representation and genomics relation analysis. Our model is fine-tuned in two stages on a comprehensive corpus composed of over 3,000,000 terms in genomics, proteomics, and medical genetics, derived from multiple large-scale validated datasets and scientific publications. GP-GPT demonstrates proficiency in accurately retrieving medical genetics information and performing common genomics analysis tasks, such as genomics information retrieval and relationship determination. Comparative experiments across domain-specific tasks reveal that GP-GPT outperforms state-of-the-art LLMs, including Llama2, Llama3 and GPT-4. These results highlight GP-GPT's potential to enhance genetic disease relation research and facilitate accurate and efficient analysis in the fields of genomics and medical genetics. Our investigation demonstrated the subtle changes of bio-factor entities' representations in the GP-GPT, which suggested the opportunities for the application of LLMs to advancing gene-phenotype research.
\end{abstract}

\section{Introduction}





The relationship between genes, phenotypes and diseases is of fundamental importance but has not yet been fully understood. Due to the complex interplays and mutual regulations among genes, proteins, metabolomics, and phenotypes, finding the proper representations of their relationship has become a significant problem. There are numerous studies that focus on single-molecule or dual-molecule biological levels such as gene mutation, proteomic, gene expression, transcriptional regulation, pathway analysis, and clinical observation. These studies aim to address the questions in specific areas using simplified modelling. Existing research in human populations has elucidated the landscape of complex genomic relations. For instance, Genome-wide association study (GWAS)~\cite{visscher201710, watanabe2019global, jansen2019genome} is a well-established and effective approach for statistically discovering genetic loci associated with common diseases. Meta-analysis of genetic association provides new insights into cohorts of independent gene–disease associations study by combining them and resolving the discrepancies~\cite{macarthur2017new}. Genome annotation aims at understanding specific functional and structural properties which were conferred by nucleotide regions and amino acid sequences~\cite{abril2019genome}. The Gene Ontology (GO) knowledge base (http://geneontology.org) comprehensively mapped the gene, functions of genes, and gene products into a formatted network~\cite{gene2023gene} while human-specific gene networks like HumanNet v2 (http://www.inetbio.org/humannet) comprises a hierarchy of networks about gene-disease-drug associations~\cite{hwang2019humannet}. The Online Mendelian Inheritance in Man(OMIM)~\cite{amberger2015omim} and DisGeNET~\cite{pinero2020disgenet} are essential knowledge platforms for disease genomics. Such large-scale genomics datasets and relevant studies have significantly expanded the scope of representative analysis of genes and diseases. However, holistic modeling at the whole genome-system scale remains challenging.

To better address this fundamental problem of representing and mapping genes and related phenotypes/diseases, the ideal approach should consider the entire genome system collectively across various molecular and biological levels. Additionally, it should incorporate experimental knowledge, bio-computational evidence, and biological text data, which have been overlooked in previous studies despite their considerable value. Large language models(LLMs) have emerged as transformative tools in natural language processing, demonstrating unprecedented capabilities in understanding and generating human-like text~\cite{raffel2020exploring, achiam2023gpt, touvron2023llama,touvron2023llama}. These models, trained on extensive corpora, have shown remarkable versatility across diverse domains~\cite{liu2023summary,zhao2023brain,ma2024iterative,dai2023chataug,liu2023radiology,liao2023differentiating,liu2023context}. There are also significant successes in applying language models to the general biomedical field~\cite{zhang2023biomedgpt,dai2023ad,holmes2023evaluating,rezayi2022clinicalradiobert,wang2023prompt,liu2023holistic}. LLM's potential to process and analyze complex, unstructured data presents a unique opportunity to address long-standing challenges in bioinformatics. The ability of LLMs to capture intricate relationships and context from text aligns well with the complexities inherent in genetic and phenotype data. This synergy opens new avenues for knowledge discovery, hypothesis generation, and the elucidation of previously obscure gene-disease associations.

In this study, we focus on transforming the aforementioned problem into an AI-based system based on a gene-phenotype large language model, named GP-GPT. The main goal of GP-GPT is to integrate multiple sources of structured and unstructured genomic knowledge into a general LLM framework. Particularly, the model leverages both structured and unstructured data from multiple main data sources from OMIM, DisGeNET, and dbGaP~\cite{mailman2007ncbi}. By converting the data into bio-text, we identified and categorized gene entities, gene functions, protein entities, protein functions, phenotype entities, genotype-phenotype association analysis, and related biological mechanisms. These components were then integrated into an informative context, which was subsequently used to develop the GP-GPT in the form of generative language models. Therefore, the model considers multiple levels of bio-factors and achieves knowledge mapping of the entire genome system rather than individual relationships.

For the first time, as far as we know, we applied large language models on multi-levels of genomics data. Our study focuses on three critical NLP tasks in the domain of genetic and phenotypic data analysis. First, we explore the model's capacity for question answering, assessing its ability to provide accurate responses to complex queries in medical genetics. 
Second, we assessed the model's performance in retrieving genomics information, particularly its ability to identify associated genes or phenotypes by providing the corresponding information.
Finally, we examined the model's proficiency in relation determinations and evaluated its ability to accurately identify relationships between genotypes and phenotypes in given contexts. These tasks are designed to evaluate the potential of language models in enhancing our understanding and interpretation of genomics data, potentially accelerating discoveries in the field of medical genetics. We observed a better performance of GP-GPT on multiple metrics against benchmark language models. Thus, GP-GPT enables simplified and quick retrieval of crucial medical genetics information (e.g., molecule name, functions, interactions, and disease associations). 

In addition, GP-GPT indirectly encodes the genomics knowledge graph. Especially for the phenotype/disease gene mapping, the model extended beyond the generation of trustworthy answers based on literature-based genomics questions provided by users. Interestingly, we observed the distribution of the gene embeddings and phenotype/disease embeddings. Compared to benchmark language models, our GP-GPT encodes genomics entities more efficiently in terms of both model depth and training progression. This aggregation and balancing of latent embeddings could be observed in terms of gene-phenotype relation pairs and gene-disease tissue distribution. The results show that the learned representations of genomics information in GP-GPT have the potential to support the holistic human genomics knowledge graph and play as a strong knowledge-aware system to provide prior probability in quantitative association analysis.

\section{Related Work}
\subsection{Language models for Knowledge Embedding and Relation Extraction}
Recently, language models such as ChatGPT and GPT-4~\cite{achiam2023gpt}, have shown remarkable capabilities in accelerating research innovation across natural sciences. ChatGPT, GPT-4, GPT-4V, and GPT-4 Turbo, developed by OpenAI, are prominent language models based on the autoregressive transformer decoder architecture, which utilizes self-attention mechanisms. These models, trained on extensive text corpora, excel in generating human-like text and demonstrate profound comprehension across diverse knowledge domains. In contrast, open-source models like Meta’s LLaMA~\cite{touvron2023llama} and Llama2~\cite{touvron2023llama}, which feature up to 70B parameters in their largest versions, aim to enhance training efficiency and reduce computational costs. Similarly, BLOOM (176B version)~\cite{le2023bloom} and BLOOMZ (176B version)~\cite{muennighoff2022crosslingual} emphasize linguistic versatility, highlighting a commitment to openness and accessibility. The more recent Falcon~\cite{almazrouei2023falcon}, with 40B parameters in its largest configuration, has also shown excellent performance on various benchmarks. These models exemplify the cutting-edge of universal language model research. LLaMA models, in particular, have performed exceptionally well on diverse benchmarks, making them some of the most popular open language models to date, hence their selection for analysis in this paper. 

Biomedical natural language processing enables the automatic extraction of crucial information from medical literature, including insights into genetic diseases and associated variants. A key task in this field is Biological Relation Extraction (RE), which involves identifying relationships between two or more entities within specific contexts, which is crucial to identify biological entities such as genes, proteins, diseases, drugs, and miRNAs in textual data~\cite{gill2024large,pinero2020disgenet,babbi2017edgar,marchesin2022tbga}. Biological RE models are widely used to extract knowledge from various sources, including medical literature. A key task in Biological RE is identifying relationships between two or more entities within specific contexts. To advance NLP and machine learning techniques for biomedical RE, considerable effort has been invested in developing relevant corpora~\cite{kim2003genia, pyysalo2007bioinfer, krallinger2008overview, luo2022biored}. Unlike simple regular expression-based methods, Biological RE models leverage the entire context of a sentence to accurately identify and classify these entities, overcoming the limitations of more traditional approaches. Although machine learning (ML) techniques generally outperform traditional methods, they require substantial amounts of manually annotated training data. This presents a significant challenge, especially in the genomics domain, where labelling training sequences at the individual word level is time-consuming and requires domain-specific expertise. Owing to large language models which offers a solution by enabling fine-tuning for Biological RE tasks with minimal training data, thereby reducing the burden of extensive manual annotation.

\subsection{Instruction Fine-tuning and Parameter efficient Training}
Instruction fine-tuning has emerged as a pivotal technique in improving the performance and generalization of language models across various tasks. The work by Wei et al.~\cite{wu2023interpretability} introduces a novel training methodology that leverages instruction fine-tuning to enhance model capabilities. By fine-tuning models on multiple tasks with carefully curated instructions, this approach not only improves task-specific performance but also enhances the model's ability to generalize across different domains.

The Stanford CRFM report~\cite{stanfordStanfordCRFM} further substantiates the efficacy of instruction fine-tuning. The report provides a comprehensive evaluation of different models and training strategies, demonstrating that instruction fine-tuning can achieve significant improvements in performance even with limited computational resources. This underscores the potential of instruction fine-tuning as a cost-effective solution for developing versatile language models.

Parameter-efficient training methods aim to optimize model performance while minimizing computational overhead~\cite{hu2021lora}. Recent advancements in this field are well-documented in the work by Liu et al.~\cite{stanfordStanfordCRFM}, which explores various parameter-efficient techniques~\cite{houlsby2019parameter}. These techniques are designed to reduce the number of trainable parameters without compromising model accuracy, making them particularly suitable for environments with constrained resources .

Key innovations include low-rank adaptation, pruning, and quantization, which collectively contribute to maintaining model efficacy with fewer parameters. Experimental results in Liu et al.~\cite{ouyang2022training} highlight that such methods can achieve near state-of-the-art performance with a fraction of the computational cost, thereby promoting sustainable AI development practices.

Combining instruction fine-tuning with parameter-efficient training presents a promising avenue for future research~\cite{dettmers2024qlora}. Integrating these methods can potentially lead to the development of highly adaptable and resource-efficient models. The synergy between these approaches lies in their complementary strengths: instruction fine-tuning enhances task adaptability, while parameter-efficient training ensures computational feasibility.

Overall, the convergence of instruction fine-tuning and parameter-efficient training methodologies signifies a crucial step towards creating robust, efficient, and scalable AI systems. Continued exploration and refinement in these areas are expected to drive further innovations in the field of machine learning and artificial intelligence.

\subsection{Language Model Applications in Bioinformatics}
In recent years, the emergence of language models for bioinformatics has shown significant progress, with notable examples including AlphaFold2~\cite{bryant2022improved}, AlphaFold3 ~\cite{abramson2024accurate}, GeneGPT~\cite{jin2024genegpt}, BioT5~\cite{pei2023biot5}, BioT5+~\cite{pei2024biot5+}, and Med-Gemini~\cite{yang2024advancing}. These models can be categorized into single-domain and multi-domain approaches, each leveraging different strategies for pretraining and application in the field.

Single-domain approaches pretrain their models solely on tokens from a specific domain. A prominent example is GeneGPT~\cite{jin2024genegpt}, which introduces a novel method for teaching large language models (LLMs) to utilize the Web APIs of the National Center for Biotechnology Information (NCBI) to answer genomics-related questions. By focusing exclusively on genomics data, GeneGPT can provide precise and relevant responses tailored to this domain.

Based on different foundational models, large language models in the bio-informatics domain can be categorized into BERT-based LLMs, T5, GPT and others~\cite{bi2024ai}. BERT-based large models related to biology include the SciBERT model. SciBERT is pre-trained on scientific publications from Semantic Scholar, mainly using the WordPiece unsupervised tokenization technique~\cite{beltagy2019scibert}.
ClinicalBERT is a pre-trained model built on a BERT-based architecture using clinical notes as training data. The types of clinical notes include nursing, radiology, and others~\cite{alsentzer2019publicly}. 
SciFive is a T5-based generative model pre-trained on millions of PubMed abstracts and PMC scientific publications. This model excels at handling biomedical tasks such as question answering and named entity recognition, outperforming models like BioBERT, T5, and BERT~\cite{phan2021scifive}.
BioGPT is a GPT-2-based model pre-trained on 15 million PubMed articles, making it one of the first GPT-inspired biomedical frameworks. At the time, it was also one of the top-performing models in its domain~\cite{luo2022biogpt}.

In the application of large language models to biological sequence data, research can be broadly categorized into four areas: DNA sequences, RNA sequences, protein sequences, and multi-omics sequencing data. Each of these areas presents unique challenges and opportunities for decoding and understanding biological systems.
For DNA sequences, models such as Enformer~\cite{avsec2021effective}, Nucleotide Transformer~\cite{dalla2023nucleotide}, GenSLMs~\cite{zvyagin2023genslms}, and HyenaDNA~\cite{nguyen2306hyenadna} have been developed to decipher the genetic blueprint of life, advancing our understanding of gene regulation, function, and expression.
Similarly, in the study of ribonucleic acid (RNA), foundational models like RNABERT~\cite{akiyama2022informative}, RNA-FM~\cite{chen2022interpretable} are instrumental in unravelling the complex mechanisms that drive RNA-mediated biological processes. These models play a crucial role in areas such as transcriptomics and gene expression analysis.
In the protein domain, where the focus is on protein structure, function, and interactions, numerous models have been proposed, such as AlphaFold2~\cite{bryant2022improved}. The model aims to predict protein folding, dynamics, and interactions, all of which are essential for understanding cellular processes and developing therapeutic strategies. 
Multi-omics sequencing data integrates various biological layers and enables the analysis of more complex and comprehensive downstream tasks. Models like Geneformer ~\cite{theodoris2023transfer}, scFoundation ~\cite{hao2024large}, AlphaFold3~\cite{abramson2024accurate}, and scGPT ~\cite{cui2024scgpt} is at the forefront of this domain, offering powerful tools for integrating single-cell data and other omics modalities to gain a holistic view of biological systems.

On the other hand, multi-domain approaches employ a diverse array of data modalities during pretraining. BioT5+, for instance, utilizes molecular structures, textual descriptions, and molecular naming conventions~\cite{pei2024biot5+}. This approach enhances the model's ability to perform a wider range of downstream tasks. Med-Gemini extends this concept further by incorporating an even broader spectrum of multimodal data, including 2D and 3D radiology images, histopathology patches, ophthalmology images, dermatology images, and genetic data~\cite{yang2024advancing}. This extensive pretraining allows Med-Gemini to address a vast array of biomedical tasks.

Furthermore, these models often function through an interactive interface, providing chatbot-like functionalities for downstream tasks. Models like BioT5~\cite{pei2023biot5} enable users to engage in intuitive dialogue formats, posing inquiries in plain text and receiving detailed, contextually relevant responses. This conversational ability allows chemical LLMs to perform complex tasks that require understanding and reasoning beyond simple query responses.
The majority of these models achieve their functionality through two primary methods: fine-tuning on domain-specific data or directly interfacing with existing APIs to provide responses. For example, GeneGPT~\cite{jin2024genegpt} uses carefully designed prompts to demonstrate to the LLM how to query APIs effectively, utilizing appropriate APIs during the inference phase to respond to user requests. Alternatively, models like Med-Gemini can directly answer user queries through the model’s outputs, demonstrating versatility and a comprehensive understanding of a wide range of topics~\cite{yang2024advancing}.

\subsection{Mapping of Genotypes and Phenotypes}
Genotype-to-phenotype mapping is a traditional work in bioinformatics. In a statistical view, GWAS~\cite{visscher201710, watanabe2019global, jansen2019genome} have played a dominant role in discovering potential relations between genotypes and phenotypes on a population scale. It has tested hundreds of thousands of genetic variants across human and other genomes to find those statistically associated traits or diseases, through which gaining insight into a phenotype’s underlying biology, estimating its heritability, and discovering genetic correlations. From a micro view, advancements in sequencing technologies, particularly single-cell RNA-seq (scRNA-seq), have significantly advanced our understanding. These techniques expose the intricate dynamics of gene expression at the cellular level, unveiling the extensive landscape of genotypes shaped by numerous factors~\cite{norman2019exploring}. In recent years, the utilization of machine learning, particularly the adoption of self-supervised transformer models originally developed for natural language processing (NLP), has demonstrated significant potential in the analysis of complex biological datasets ~\cite{theodoris2023transfer, cui2024scgpt}.

At the cohort level, meta-analysis is an effective approach to integrate multiple association studies of genes and phenotypes. Meta-analysis is a statistical tool used to combine results from different studies on the same topic and quantitative synthesis of research findings. Meta-analysis enables the identification of genuine associations between genes and phenotype~\cite{trikalinos2008meta}. Particularly for the studies based on human populations, a comprehensive catalogue of all published GWAS and association results have become publicly available, which provided resources for human-specific meta-analysis~\cite{macarthur2017new}.

At the level of human cells, there is a principal goal of revealing the landscape of a combination of genes that results in an extraordinarily diverse phenomenon. Phenomic databases~\cite{freimer2003human}, have been collected as comprehensively as systematically assemblages of phenotype information.\textit{Freimer, et. al.}~\cite{khodaee2024multimodal} has presented a computational integrated genetics framework designed to analyze and interpret the high-dimensional landscape of genotypes and their associated phenotype. The study introduces the use of self-supervised language models to simultaneously map the genotype-phenotype landscape, proposing the concept of integrated genetics as an alternative to traditional forward and reverse genetics. In this approach, phenotypes such as sex, age, anatomical tissue, and cell types are integrated and learned alongside genotypic data obtained from scRNA-seq. This method deepens our understanding of the biological context of gene expression and, ultimately, the genotype-phenotype relationship.

\subsection{Gene Networks and Knowledge Graphs}
Whole genome sequencing~\cite{cho2016muffinn, horn2018netsig} and GWAS studies~\cite{greene2015understanding, shim2017gwab}have already provided substantial evidence for investigating disease-associated genes. Recent research suggests that genomics datasets can be represented as gene networks, where the nodes symbolize genes, phenotypes, or proteins, and the connections between the nodes indicate the relationships between these entities. Schlitt, et.al. ~\cite{schlitt2003gene} propose a novel method for identifying functionally related genes by comparing neighbourhoods within gene networks. This approach does not depend on gene sequence or protein structure homologies and can be applied to any organism and a wide range of experimental datasets. Gene networks enable researchers to seek additional evidence to cross-check their results using various sources, e.g. enriching existing computational approaches such as linkage analysis and GWAS studies by including the relationships identifications between genes from protein-protein interaction (PPI), gene expression, gene ontology and other domain knowledge ~\cite{ata2021recent}. 
HumanNet, HumanNet2, and HumanNet3 constructed genome-scale functional networks for human genes using a variety of evidence sources, including human mRNA co-expression, protein-protein interactions, protein complexes, and comparative genomics datasets. In these networks, the edges may represent different types of associations between genes, which can be mapped through both experimental and computational methods~\cite{kulikajevas2021humannet, hwang2019humannet, kim2022humannet}.

At the same time, the development of knowledge bases and graphs has greatly enhanced performance in fields like computer vision and natural language processing. Likewise, knowledge graphs have been widely applied to specific biological tasks, such as cancer diagnosis~\cite{elmarakeby2021biologically, yu2016translation, ma2018using}.
Other studies take advantage of the Gene Ontology (GO) knowledgebase to build the neural network architecture to simulate the gene or protein reactions in the cells~\cite{yu2016translation, ma2018using}. Combining multiple bio-factors, biological factor regulatory neural network (BFReg-NN) ~\cite{dai2023biological} aims at simulating the complex biological processes in a cell system, understanding the functions of genes or proteins.

\section{Method}
\subsection{Data}
The datasets in genomics, proteomics, and medical genetics have experienced significant growth in both scale and diversity. Our dataset-building attempt starts from the gene entities and their textual descriptions maintained by the National Center for Biotechnology Information (NCBI)~\cite{ncbi_url}. Taking the 192064 individual gene items from NCBI as the entries, we then collected and screened the data of related bio-factors from multiple datasets. Each of these datasets stands for a particular level in the genomics system(Fig.~\ref{fig1}).

\begin{minipage}[b]{1.02\linewidth}
  \centering
  \centerline{\includegraphics[width=\textwidth,height=8.5cm]{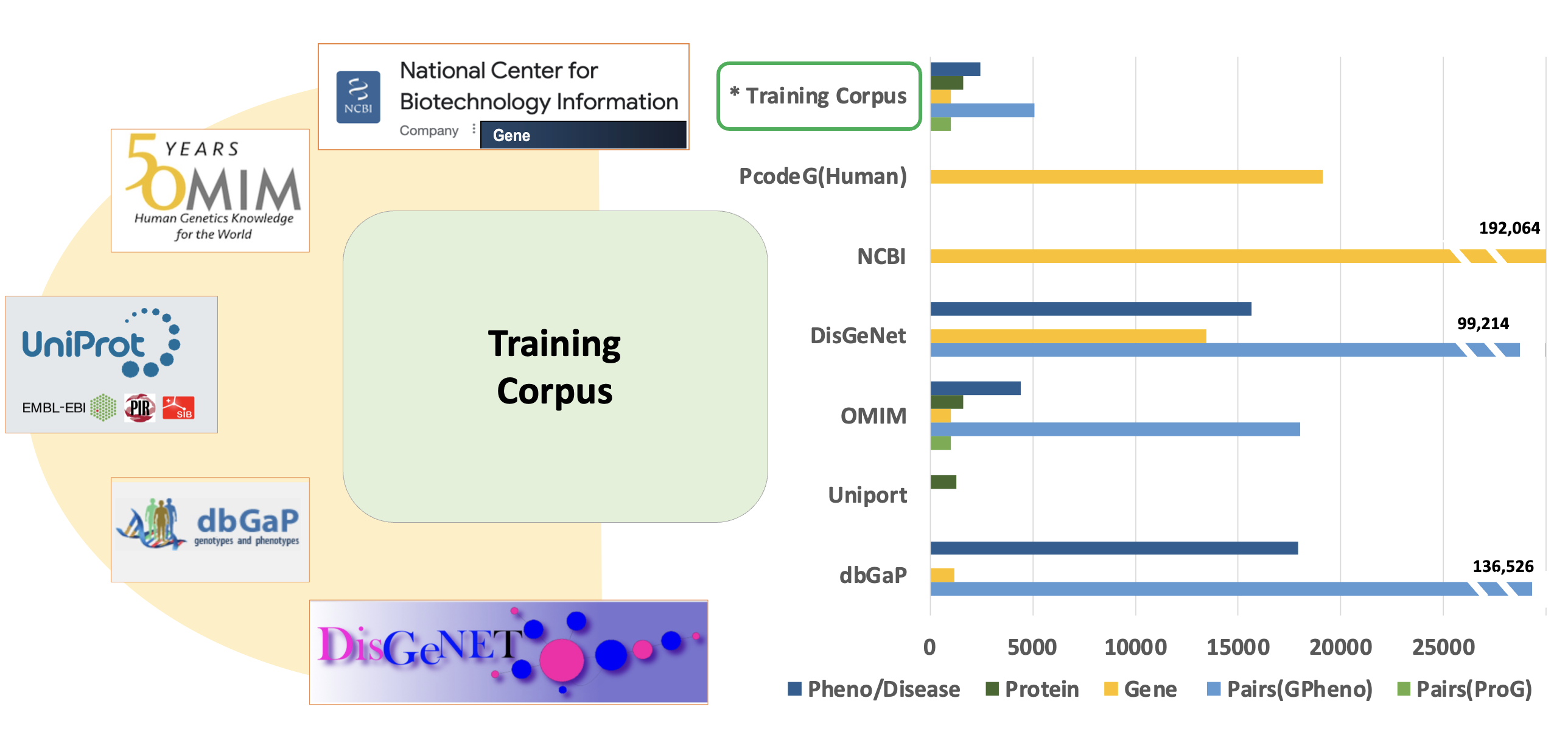}}
  \captionof{figure}{Integration of multiple datasets.
}
\label{fig1}
\end{minipage}

\subsubsection{Data Collection}

For the purpose of comprehensively incorporating and extracting the mapping between genes and associated phenotype/diseases, multiple data sources were integrated into the training dataset of this study. A major part of the dataset came from the Online Mendelian Inheritance in Man(OMIM) database~\cite{omim_url}, which is a comprehensive and authoritative resource for understanding human genes and their associated genetic disorders. OMIM keeps full-text information on all known Mendelian disorders and over 16,000 genes and provides detailed information on human genes, genetic phenotypes, and the relationships between genes and phenotypes. Genes often cast influences on phenotypes through gene products and a series of related complex biological pathway alterations ~\cite{amberger2015omim}. Building on this concept, we incorporated the information of gene products from UniPort~\cite{coudert2023annotation}, and their molecular function information from NCBI. The extracted pairs of genes and associated phenotypes/diseases were later verified and supplied through dbGaP~\cite{mailman2007ncbi} and DisGeNet~\cite{pinero2020disgenet}.
In summary, from OMIM, we collected the gene-phenotype pairs along with the relational graphs. A total of 4401 pairs of phenotype and gene have been collected and verified with the dbGaP data. From UniPort, we collected 1286 gene-protein pairs. From DisGeNet, we use the Relation Extraction (RE) samples from the evidence terms. We integrated these multi-level genomics text data to serve our training objective. All the mentioned data are examined and matched to the NCBI gene IDs and organized accordingly.

\begin{minipage}[!b]{1.0\linewidth}
  \centering
  \centerline{\includegraphics[width=\textwidth,height=9.2cm]{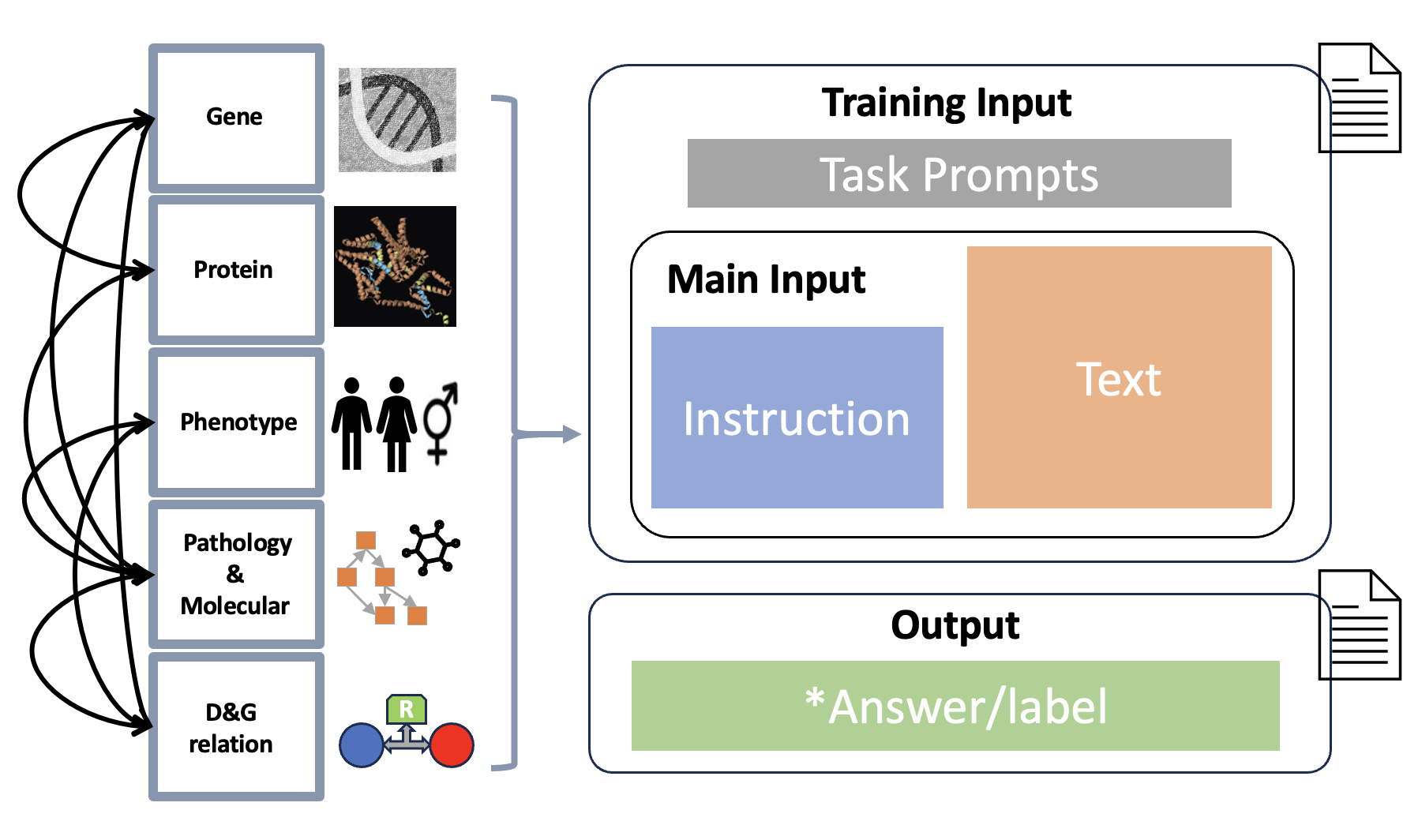}}
  \captionof{figure}{Overview of multi-task multi-level formats of genomics text data. The training data set can be built on intrinsic logic in multiple genomics datasets.}
  \label{fig2}
\end{minipage}

\subsubsection{Construction of the Multi-task and Multi-level Genomics Training Corpus}
Based on the intrinsic logic in multiple data sets, we extracted relevant bio-text and structured data and formulated them into a multi-task multi-level genomics training corpus. The corpus is composed of three training datasets, each containing different types of genomics knowledge context. The first dataset is designed for the initial training stage, which follows the masked training manner, while the last two datasets are designed for conventional instruction fine-tuning.

The first part of the training set is based on instruction mask training. In this dataset, gene entities data and phenotype/disease entities data from the dbGaP are organized into masked text data (Fig.~\ref{fig3}). The model is trained to follow the instructions to properly predict masked keywords. The structure of the input text data follows the standard format of instruction fine-tuning guidance provided by Llama models.

The second part of the training set focuses on excavated gene–protein relationships (Fig.~\ref{fig4}). In this dataset, named gene-protein contexts, we group the genomics knowledge based on three types of relations: \textbf{Gene-protein relation} formulated the gene entity, the protein entity, the gene functions, and the protein functions together. This format contains two sub-formats. The first sub-format focuses on identifying the function of protein given its coding gene and gene function, while the second sub-format focuses on identifying gene products. \textbf{Protein-function inferring} formulated both protein entities and their functions. This sub-format focuses on training the model to identify the protein entity based on relevant molecular functions. \textbf{Gene-function inferring} formulated the gene entity and its functions. This sub-format focuses on training the model to identify the gene entity based on the gene's functions. Each type of the three relations is composed of text fields of 'Task Prompt', 'Input', and 'Output'. 

The third part of the training set is based on the triple relationships between gene, protein, and disease/phenotype entities (Fig.~\ref{fig5}). This dataset is a gene-protein-phenotype/disease contexts dataset and is structured based on the relations within this trinity: gene, protein, and disease/phenotype. By considering the protein entities as a pivot, the dataset attempts to capture the complex relations between these entities by examining their molecular and functional relations with the protein entities, thereby, guiding the model to learn such relationships. To be noted that the three kinds of contexts in this dataset contain both positive pairs and artificial negative pairs. Following the aforementioned format, each context follows a formula of 'Task Prompt', 'Input', and 'Output'. \textbf{Protein-molecular features} formulated the protein entity, the phenotype, and the molecular relations, focusing on identifying the relationship between protein and phenotype/disease while providing the underlying molecular mechanism. \textbf{Protein-pathogenesis features} formulated the protein entity, the phenotype, and the pathogenesis relations, focusing on identifying the relationship between protein and phenotype/disease by providing bio-pathway information. \textbf{Gene-inheritance features} formulated the gene entity and inheritance facts. This format focuses on deciding the gene entity given the relevant inherited facts of diseases.

\begin{minipage}[b]{1.01\linewidth}
  \centering
\centerline{\includegraphics[width=\textwidth,height=20.2cm]{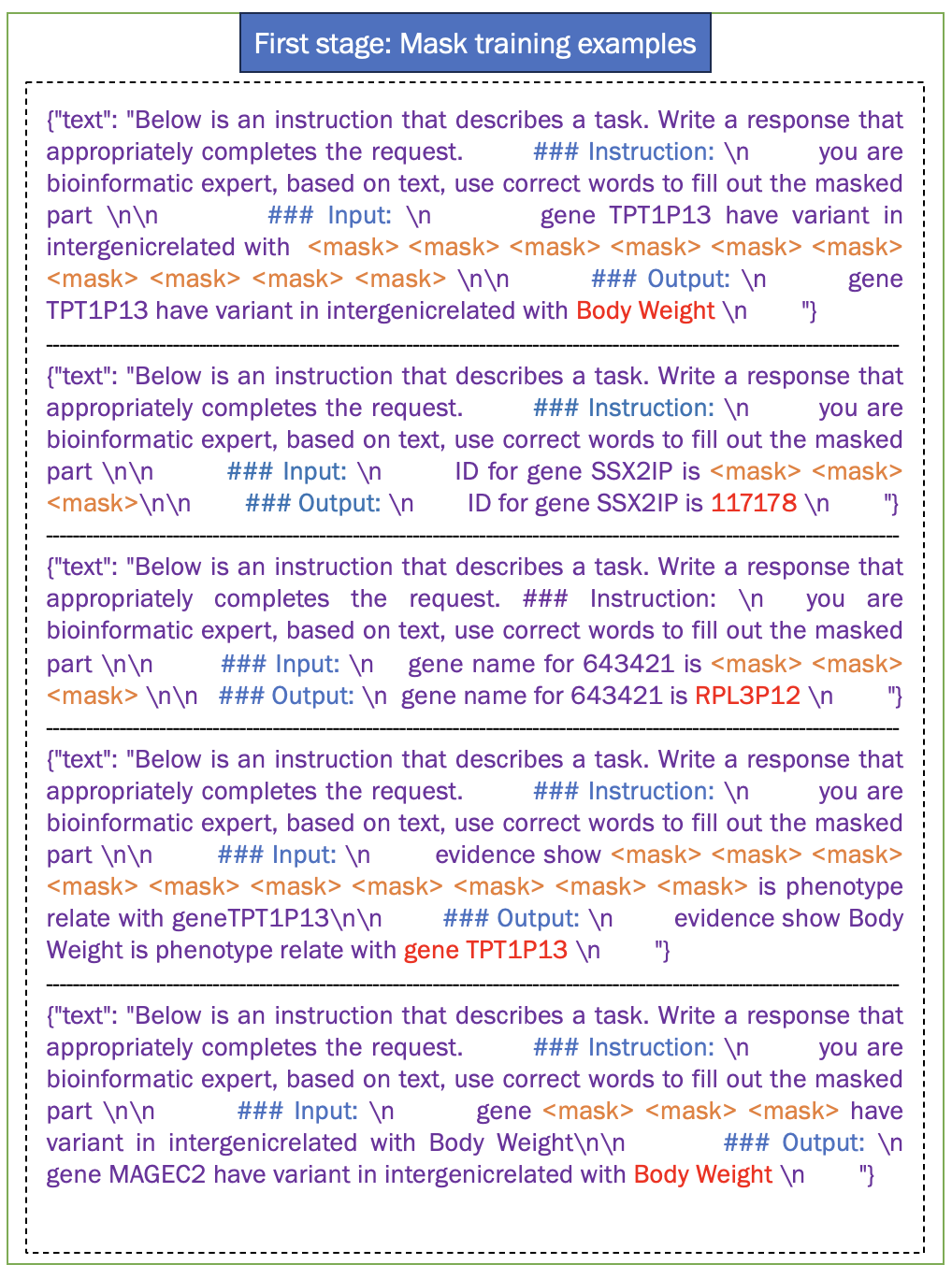}}
  \captionof{figure}{Tuning model at the first stage using instruction mask training data. The bio-text has been fitted into the input format provided by the Llama model. The signs: '\#\#\# Instruction:, '\#\#\# Input:', and '\#\#\# Output:', stand for the indicators inside the model input. The red words indicate the replaceable gene entities and phenotype entities.}
  \label{fig3}
\end{minipage}

\begin{minipage}[b]{1.02\linewidth}
  \centering
\centerline{\includegraphics[width=\textwidth,height=18.5cm]{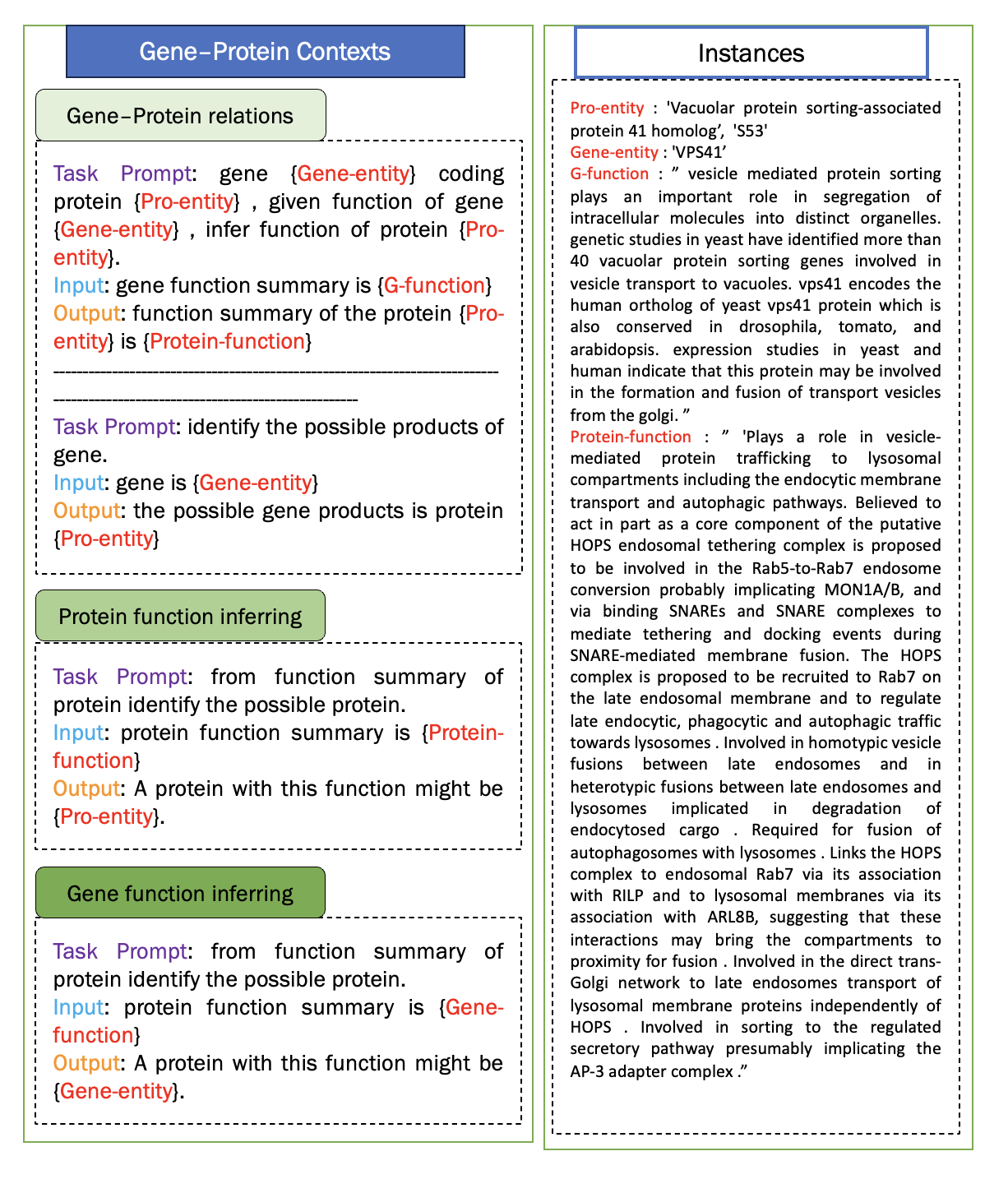}}
  \captionof{figure}{Formatted genomics contexts of gene–protein. The red words indicate the replaceable gene entities and phenotype entities. The examples in the 'Instances' column show the available text which can be used to fit into the position of red words in the Gene-Protein contexts.}
  \label{fig4}
\end{minipage}

\begin{minipage}[b]{1.02\linewidth}
  \centering
\centerline{\includegraphics[width=\textwidth,height=18.5cm]{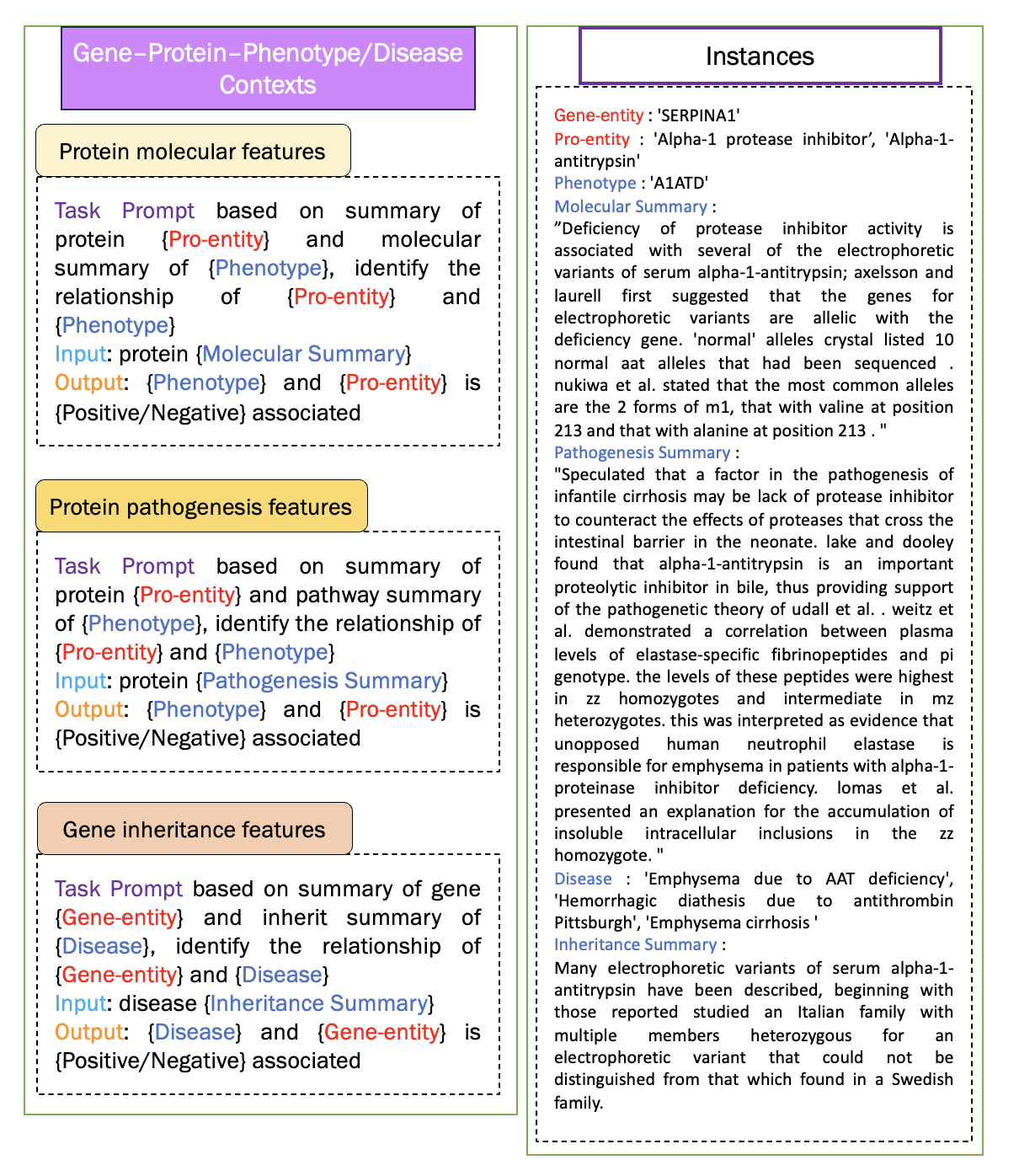}}
  \captionof{figure}{Formatted genomics contexts of gene-protein-disease/phenotype. The red words indicate the replaceable gene entities and phenotype entities. The examples in the 'Instances' column show the available text which can be used to fit into the position of red words in the Gene-Protein-Phenotype/Disease contexts.}
  \label{fig5}
\end{minipage}

\subsection{Model}
In this section, we introduce GP-GPT, an LLM-based paradigm developed to precisely map gene and phenotype entities using bio-text linguistic knowledge. GP-GPT uses the Llama2 framework to establish an autonomous LLM-based framework with a prompt question-and-answer format. This model can automatically search for latent gene-phenotype associations, extract meaningful relations, summarise bio-function descriptions, identify bio-factors, and visualize the embedding of these bio-factors. The overview of the training input of the GP-GPT framework is shown in (Fig.~\ref{fig2}). We have constructed a training corpus with selected human genomics texts on multiple bio-system levels following the conventions in section 3.1. The text data in this corpus reflects the hidden connection between bio-factors from different levels and further organizes different types of such relations into a similar format by choosing different instructions.

\subsubsection{Parameter Efficient Fine-tuning}

In this study, we utilize the parameter-efficient fine-tuning techniques of the LLaMA model family (Llama2, Llama3, Llama3.1) using Low-Rank Adaptation (LoRA)~\cite{hu2021lora}, and QLoRA~\cite{dettmers2024qlora} technique. The objective is to adapt the Llama3 model for specific downstream tasks while minimizing the number of trainable parameters and computational resources required. LoRA introduces a low-rank adaptation matrix into the transformer architecture, allowing efficient fine-tuning with fewer parameters. In addition, QLoRA applies a quantized pre-trained language model into low-rank adapters, significantly reducing memory usage to further finetune large language models. Both techniques modify the original model minimally, making it computationally efficient and less prone to overfitting, especially with limited data. In our study, we tested both LoRA and 8-bit QLoRA on the Llama2 7B model and the Llama3.1 8B model. With no significant performance discrepancies between the two techniques, we employed 8-bit QLoRA to achieve efficient fine-tuning. Additionally, 4-bit QloRA was used during the fine-tuning of the Llama3.1 70B model.

We built the fine-tuning models on the pre-trained Llama2 model 7B, Llama3.1 model 8B, and eventually on the pre-trained Llama3.1 70B. LoRA layers are inserted into the transformer architecture of such  Llama models. These layers are designed to learn task-specific adaptations by introducing low-rank matrices into the attention heads of the transformer model. By keeping the base model parameters frozen, only the parameters of the LoRA layers are updated during the fine-tuning process. This approach significantly reduces the number of trainable parameters compared to traditional fine-tuning methods.

\subsubsection{Training on Genomics Data}
With the multi-task muti-modality genomics training corpus at hand, we follow the workflow of parameter-efficient fine-tuning to train GP-GPT. As shown in (Fig.~\ref{fig6}), the multi-level genomics training data have been formatted into training contexts and fed into an instruction-fine-tuning process based on Llama2. The whole instruction fine-tuning process can be viewed as two-stage training. During the first stage, the training text data is from the first part of the training set, which is prepared in the form of instruction mask prediction fine-tuning. In the second fine-tuning stage, the training text data are drawn from the second part of the training set, which is designed for supervised fine-tuning in a question-answer format.

\textbf{Tokenization and Embedding:} Models such as Llama, Llama2 and Llama3 utilize a character-based approach to numerical tokenization, where each digit and special character is treated as a separate token. This technique has been proven to be more effective in a variety of arithmetic tasks. Similarly, GP-GPT adopts this character-based tokenization method for bio-factor entities, including gene IDs, protein names, and disease abbreviations. The original dictionary remains unaltered. We assigned the <mask> token back to pre-trained Llama models during the first stage of fine-tuning, in order to implement instruction mask fine-tuning. Specifically, the re-added <mask> token has been assigned to the vocabulary dictionary at ID 35073.

\textbf{Model architecture:} GP-GPT employs the same architecture as Llama family models, with a specific LoRA adaptor merged with the original model. We follow the configuration used in Llama2(7B), Llama3.1(8B), and Llama3.1(70B). The vocabulary size of GP-GPT is 35, 073, the same as the default configuration. GP-GPT model inherits M parameters from the original Llma2(7B) and comprises an additional LoRA adaptor of M.

\textbf{Fine-tuning:} The entire instruction fine-tuning process can be considered as a two-stage training approach. In the first stage, the training content is drawn from the initial segment of the training set, which is prepared for fine-tuning through instruction mask prediction. In the second stage, the training content is taken from the latter part of the training set, which is structured for supervised fine-tuning using a question-answer format. GP-GPT can be fine-tuned on various downstream tasks involving molecules, proteins, and text. In both stages, to bridge the gap between pre-training and fine-tuning stages and unify different genomic downstream tasks, we employ a prompt-based fine-tuning approach that converts various task formats into a sequence generation format~\cite{brown2020language, gaoetal2021making}.

\begin{minipage}[b]{0.95\linewidth}
  \centering
\centerline{\includegraphics[width=\textwidth,height=11.8cm]{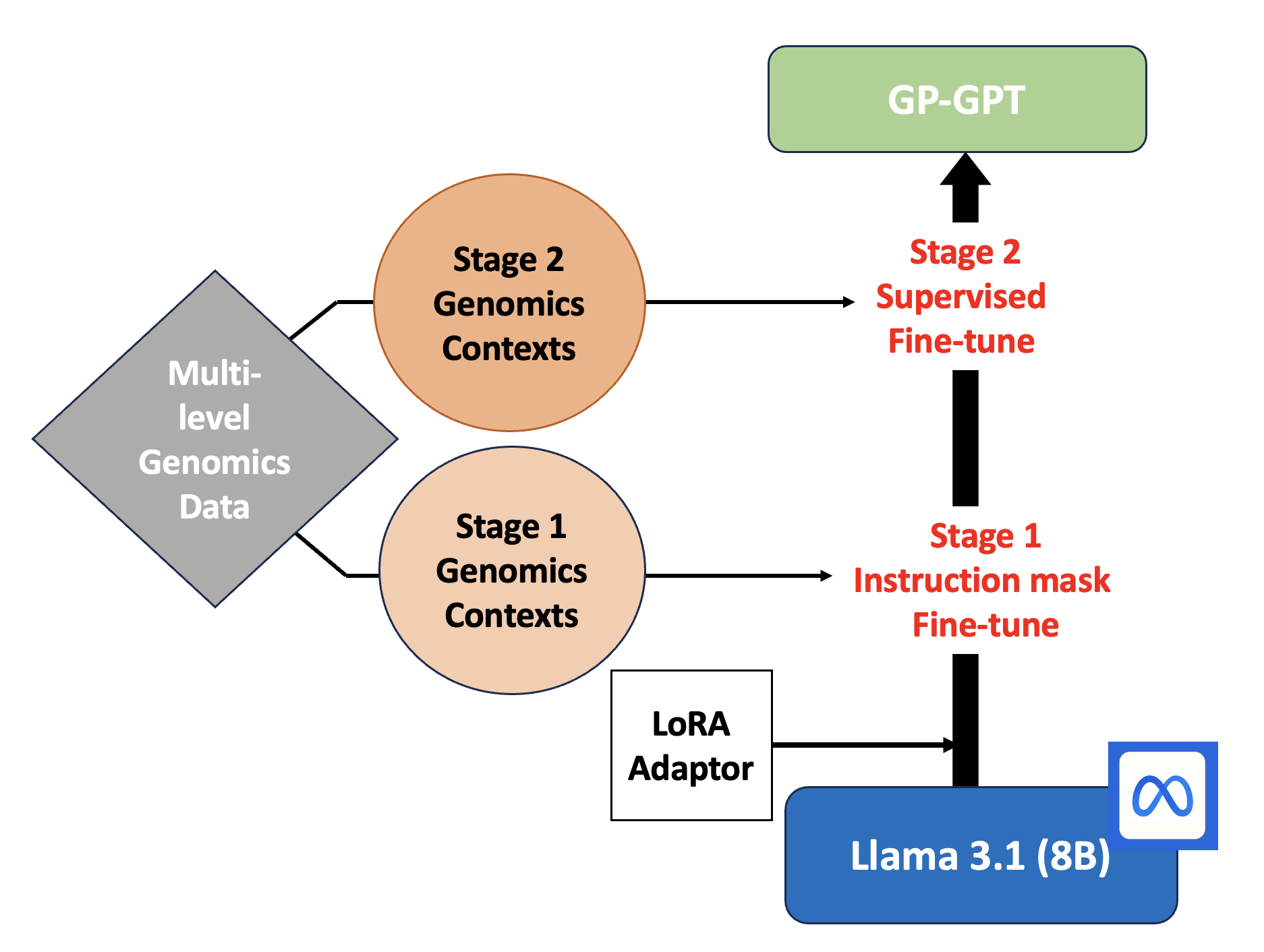}}
  \captionof{figure}{Overview of GP-GPT's training framework.}
  \label{fig6}
\end{minipage}

\subsection{Evaluation}

This study is designed to uncover the relationship between genes and phenotypes/diseases and their mutual mappings, by using linguistic information extracted from bio-text data. Therefore, considering the nature of the generative language model, we designed the evaluation experiments based on a question-answer paradigm. Besides validating the performance of the model, we are also interested in exploring the efficiency of using LLMs to map gene entities and phenotype/disease entities. We visualized a series of gene-phenotype embedding in the LLM's encoding space to better understand the potential of the genomics relation language model.

\subsubsection{Information QA}
The question-answer evaluation of gene-phenotype information in this study can be classified as a form of internal validation. Although the gene-phenotype pairs used in testing were implicitly represented in the training data, this evaluation method serves as an effective tool for assessing our model.

\textbf{Case studies:} We first evaluate GP-GPT on the text generation task. The models were required to answer the questions about gene and disease knowledge. The 'Gene disease association' questions from the QA database GeneTuring~\cite{hou2023geneturing} have been used to test our model, and an expert evaluation has been conducted by comparing the model-generated texts and golden standard answers. 

\textbf{QA evaluation:} This module tests the ability of GP-GPT models to identify the gene-disease association with generated text based on specific question-answer instruction. The ground truth of gene-disease association was downloaded from OMIM. More than 1000 gene-disease pairs were randomly selected and all genes associated with each disease were recorded. In the question-answering tasks, a question was created by pasting “What are genes related to ” before the disease name and a question mark after the disease name. To create sentence completion tasks, a question was created by pasting “The name of the gene related to ” before the disease name and “ is” after the disease name. 
The testing of disease-gene association was designed in the same style by reversing a question and answer from the above case. 
The set of gene-disease-associated pairs was used as the gold standard. To compare the model-generated answer with the gold standard, 4 metrics were applied. BLEU has been used to directly compare the two texts. The specialized BLEU-1 is designed to precisely measure the gene or disease, the shortest phrases which only contain the gene's name or disease's name are used as references in BLEU-1 calculation, while the brevity penalty and n-gram precision are neglected. 
The accuracy gene-phenotype (ACC.G-P) and the accuracy phenotype-gene (ACC.P-G) are calculated by matching the exact gene or phenotype name in the generated sentence. Note that for complex names, only the non-duplicated stem tokens in the names are considered.

We evaluate the GP-GPT against other 7 large language models. The setting of the same prompt components has been used across all different models. The candidate LLM includes: original Llama2(7B), Llama2(13B), Llama3(8B), Llama3.1(8B), Llama3.1(70B), and the Bio-GPT~\cite{luo2022biogpt}, and the GPT-4 model~\cite{achiam2023gpt}. Our evaluation of GP-GPT is more rigorous than the evaluation of LLMs and the GPT models in genomics~\cite{hou2023geneturing}, which minimized manual involvement and emphasized both exact matches and partial correctness.

\subsubsection{Genomics Relation Determination}
To evaluate the model's understanding of the relation between genes and phenotype/diseases, the well-established relation determination task has been applied to directly test the model's performance. The relation determination task was formulated as a text generation task with the proper question-answer prompt. The relation determination prompt is composed of instruction and text, models are asked to generate answers in yes/no based on the text. We use the test set from TBGA~\cite{marchesin2022tbga} data. We screened the testing data by checking the gene-disease pairs with their corresponding pairs in DisGeNet data, to ensure the knowledge correctness and testing effectiveness. We tested the GP-GPT with other large language models, for the local running models, the results were collected in a 'mixed-prompt' strategy. Thus, for GP-GPT, Llama2, Llama3, Llama3.1, and Bio-GPT, each single question-answer testing data is formulated with multiple prompts. During the results collection, the best answer was selected as the final output of the model for further metric calculations. But for commercial large language models like GPT-4, only one kind of prompt was tested due to its cost. 

\subsubsection{Bio-factors Embeddings}
Considering the gene entities and phenotype/disease entities as two kinds of bio-factors, we report the results for the soundness of the implementation of the LLM in the representation of bio-factors. To this end, we explored the mapping of genes and phenotype/disease in the embedding space of a large language model. By composing the summaries of gene entities and disease entities into sentences, we obtained the sentence embeddings of the bio-factors inside the model. The observations are intended to monitor the static embedding maps of the entities both in zero-shot mode and in a fine-tuned model. In zero-shot mode, the latent embedding was extracted from the outputs of the hidden layers. In fine-tuning models, the sentence embedding was comprised of the model's original layer outputs and the LoRA module's modifications. UMAP has been applied for the visualization of the embedding of entities.

\section{Results}

\subsection{Dataset and Model Setting}
We collected a total of 2,457,469 pieces of stage 1 genomics contextual data (1021 MiB) and 546,926 pieces of stage 2 genomics contextual data (308 MiB) for model training, out of which 10,000 randomly selected texts from each data set have been set aside as validation dataset. Both the training dataset and validation dataset are balanced in terms of the multi-level multi-task formats mentioned in 3.1. The model architecture is the same as the original Llama models: Llama2(7B), Llama3.1(8B), and Llama3.1(70B). With Q-LoRA modules added to Llama2(7B) and Llama3.1(8B) models at 8-bit, and to Llama3.1(70B) model at 4-bit. We fine-tune the model specifically focusing on optimizing a smaller set of trainable parameters (overall at 7.49 percent compared to original models), making the process faster and less resource-intensive. The learning rate was set to 1.4e-05, and the batch size wa set to 32. 
As a result, we obtained the genomics-specific fine-tuned large language models with different sizes: GP-GPT small (from Llama2 7B), GP-GPT base (from Llama3.1 8B), and GP-GPT large (from Llama3.1 70B). In the final training stage (stage 2), an early stop have been placed at the 2.75 epochs for GP-GPT small and GP-GPT base, and the early stop at 1.25 epochs for GP-GPT large.

\subsection{Experiments}
\subsubsection{Genetic Medical QA}
The model can generate reliable responses to gene-disease information queries. We provided four randomly selected cases from four different questions. Upon comparing the model-generated answers with the ground truth of gene-phenotype knowledge, experts can confirm the fluency and reasoning of the model's responses (Fig.~\ref{fig7}). The model's performance fluctuates depending on the questions. We showed both the high-quantity cases and the fewer ones for comparison. In the high-quantity cases, the GP-GPT models successfully predict the correct gene names. Although in the worst cases, the model fails to recall the mapping genes, the answer correctly recognises the disease entity and delivers paragraphs with plausible reasoning.

\begin{minipage}[b]{0.95\linewidth}
  \centering
  \centerline{\includegraphics[width=\textwidth,height=14.4cm]{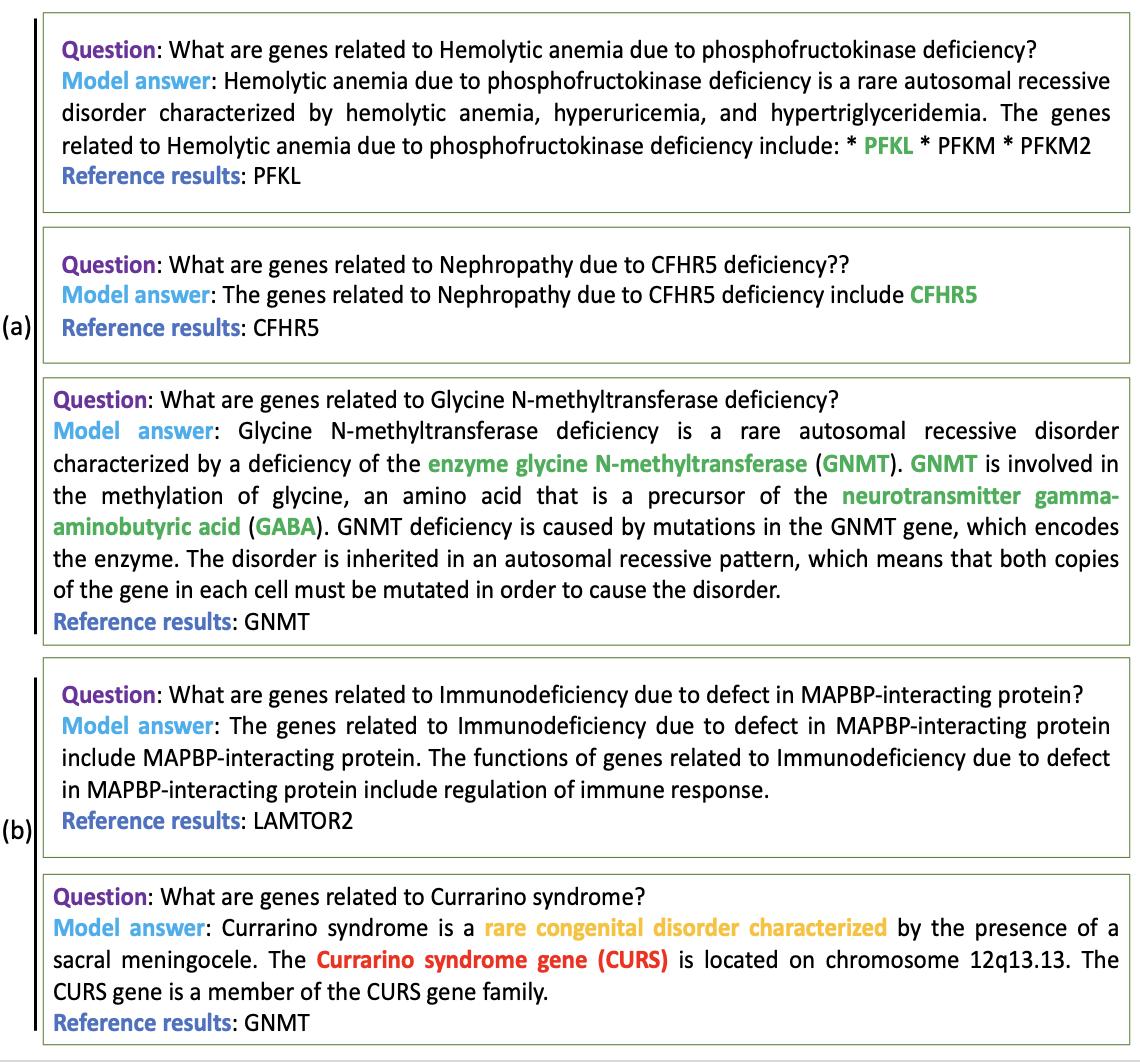}}
  \captionof{figure}{Examples of gene-phenotype mapping task on questions-answers by GP-GPT(base). (a), high quantity answers with the correct gene entity highlighted in green. (b) less desirable answers with the misleading part marked in red. the yellow phrase denotes the knowledge which not related to the question.}
  \label{fig7}
\end{minipage}

We gather 10 models to evaluate the performance of the proposed GP-GPT. The comparison results are summarized in Table.~\ref{Table1}. For GP-GPT with different sizes, their performances were tested on the four pre-defined metrics: BLEU, BLEU-1, gene-phenotype accuracy, and phenotype-gene accuracy. 
The results show that the large GP-GPT achieved the best performance on BLEU-1 and gene-phenotype/disease information recalling. When compared to commercial LLM like ChatGPT-4, GP-GPT large also demonstrates comparable performance in phenotype/disease-gene information recalling, considering their contrast in model sizes. The BLEU-1 score directly indicated the ability of a model to find correctly related gene entities by a given phenotype/disease or vice versa. These results suggest the outstanding characteristic of the GP-GPT large model in understanding domain-specific problems. While all other fine-tuned models show an adaptive ability to recall the genomics information and generate answers to the medical genetics questions, the GP-GPT base has comparable scores against other candidate models with relatively fewer parameters.

\begin{table*}[!ht]
\centering
\scalebox{1.05} {
  \begin{tabular}{@{\hspace{8pt}}ccccc@{}}
    \hline
    \textbf{Model}  &   \textbf{BLEU}  &  \textbf{BLEU-1} &  \textbf{Acc.(G-P)}  &  \textbf{Acc.(P-G)}   \\ 
    \hline
    Llama2(7B)  & 0.315 & 0.048 & 0.267 & 0.785 \\
    Llama2(13B)  & 0.260 & 0.053 & 0.288 & 0.790 \\
    Llama3(8B)  & 0.334 & 0.067 & 0.316 & 0.786 \\
    Llama3.1(8B)  & 0.318 & 0.072 & 0.322 & 0.793 \\
    Llama3.1(70B)  & 0.375 & 0.106 & 0.492 & 0.846 \\
    GPT-4 & 0.244 & 0.019 & 0.392 & \textbf{0.877} \\
    BioGPT-large & 0.298 & 0.013 & 0.270 & 0.825 \\
    \hline
    GP-GPT(base) & \textbf{0.404} & 0.102 & 0.375 & 0.847 \\
    GP-GPT(small) & 0.325 & 0.075 & 0.285 & 0.796 \\
    GP-GPT(large) & 0.323 & \textbf{0.141} & \textbf{0.533} & 0.837 \\
    \hline
  \end{tabular}
}
  \caption{
    Results of information question answer and human evaluation.
  }
  \label{Table1}
\end{table*}

\subsubsection{Evaluation on Relation Determination}
As for relation determination tasks, We gather 8 models to evaluate the performance of the proposed GP-GPT. The comparison results are summarized in Table.~\ref{Table2}. Based on the relation determination task, we evaluate GP-GPT models and other models on the four metrics: Precision, Recall, Accuracy, and F1 score.
The results show that the fine-tuning procedures have gained the ability of GP-GPT models in defining gene-phenotype relationships (based on F1 score). The overall best model is GP-GPT(small), which is based on Llama2(7B). Considering the subpar performance of original Llama3.1(8B) and Llama3.1(70B) on this special task, their corresponding fine-tuned version GP-GPT(base) and GP-GPT(large) show significant improvements and achieved comparable scores against the listed candidates.
 In summary, compared to the original models, All fine-tuned models demonstrated an adaptive ability to determine the genomics relations in the form of generating readable responses to medical genetics questions.

\begin{table*}[!ht],
\centering
\scalebox{1.15} {
  \begin{tabular}{@{\hspace{8pt}}ccccc@{}}
    \hline
    \textbf{Model}  &  \textbf{Precision}  & \textbf{Recall} & \textbf{Accuracy} & \textbf{F1}  \\
    \hline
    Llama2(7B)  & 0.303 & 0.960 & 0.310 & 0.461 \\ 
    Llama2(13B)  & 0.301 & 0.969 & 0.302 & 0.469 \\   
    Llama3(8B)  & 0.311 & 0.967 & 0.307 & 0.463 \\
    Llama3.1(8B)  & 0.315 & 0.288 & 0.305 & 0.316 \\
    Llama3.1(70B)  & 0.312 & 0.850 & 0.304 & 0.456 \\ 
    GPT-4 & 0.276 & 0.686 & 0.259 & 0.394 \\  
    Bio-bert & 0.335 & 0.530 & 0.306 & 0.411 \\ 
    \hline
    GP-GPT(base) & 0.318 & 0.537 & 0.308 & 0.400 \\
    GP-GPT(small) & 0.325 & 0.869 & 0.314 & 0.473 \\
    GP-GPT(large) & 0.313 & 0.860 & 0.307 & 0.459 \\
    \hline
  \end{tabular}
  }
  \caption{
    Evaluation on relation determination of gene-phenotype mapping.
  }
  \label{Table2}
\end{table*}

\subsubsection{Visualization of Gene-Phenotype Embeddings}
The Visualization of gene entities and phenotype/disease entities can be seen from two perspectives. First, we show the latent text representations of the gene entities and the phenotype entities (Fig.~\ref{fig8}). By comparing the distributions of sentence embeddings in the two-dimensional UMAP space, the efficiency of fine-tuning can be viewed as the distributions of gene entity embeddings and phenotype entity embeddings. In the plot, the colors indicate the IDs of the corresponding gene-disease pairs. By comparing the gene embedding plots and disease embedding plots, the embeddings from the fine-tuned model show a noticeable spreading distribution pattern, especially for diseases. 
Second, in Figure 9 (Fig.~\ref{fig9}) and Figure 10 (Fig.~\ref{fig10}), labelled by the related tissue, we probed and showed the embeddings of entities from different layers of the model in the two-dimensional UMAP space. The results suggest the representation changing of genes and diseases inside the models. Based on the comparison of GP-GPT(base) and the original pre-trained Llama model, it shows that the fine-tuning process leads to a better representation, e.g., the clustering process of neuron cell-related genes and diseases. The visualization results show that, with the training on multi-level bio-factors' data, the framework of instruction fine-tuning can effectively capture better representations of genetics biomedical entities and efficiently encode their relations.

\begin{minipage}[b]{1.05\linewidth}
  \centering
  \centerline{\includegraphics[width=\textwidth,height=15.8cm]{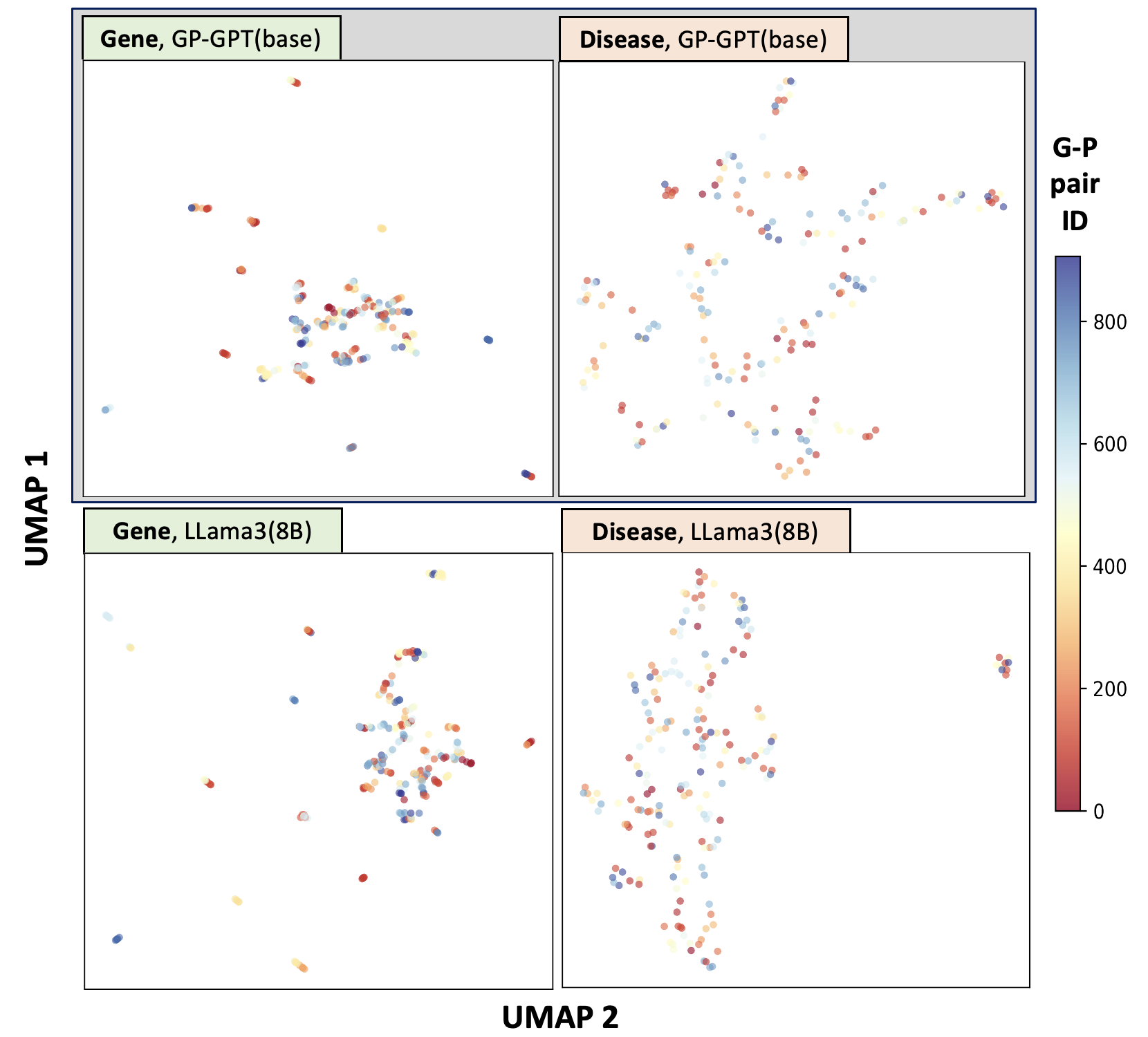}}
  \captionof{figure}{Latent text representations of gene entity and phenotype entity. Representations extracted by sentence embedding of the 30th layer from the model(GP-GPT base). Colors indicate the IDs of gene-phenotype pairs. }
  \label{fig8}
\end{minipage}

\begin{minipage}[b]{1.03\linewidth}
  \centering
  \centerline{\includegraphics[width=\textwidth,height=18.8cm]{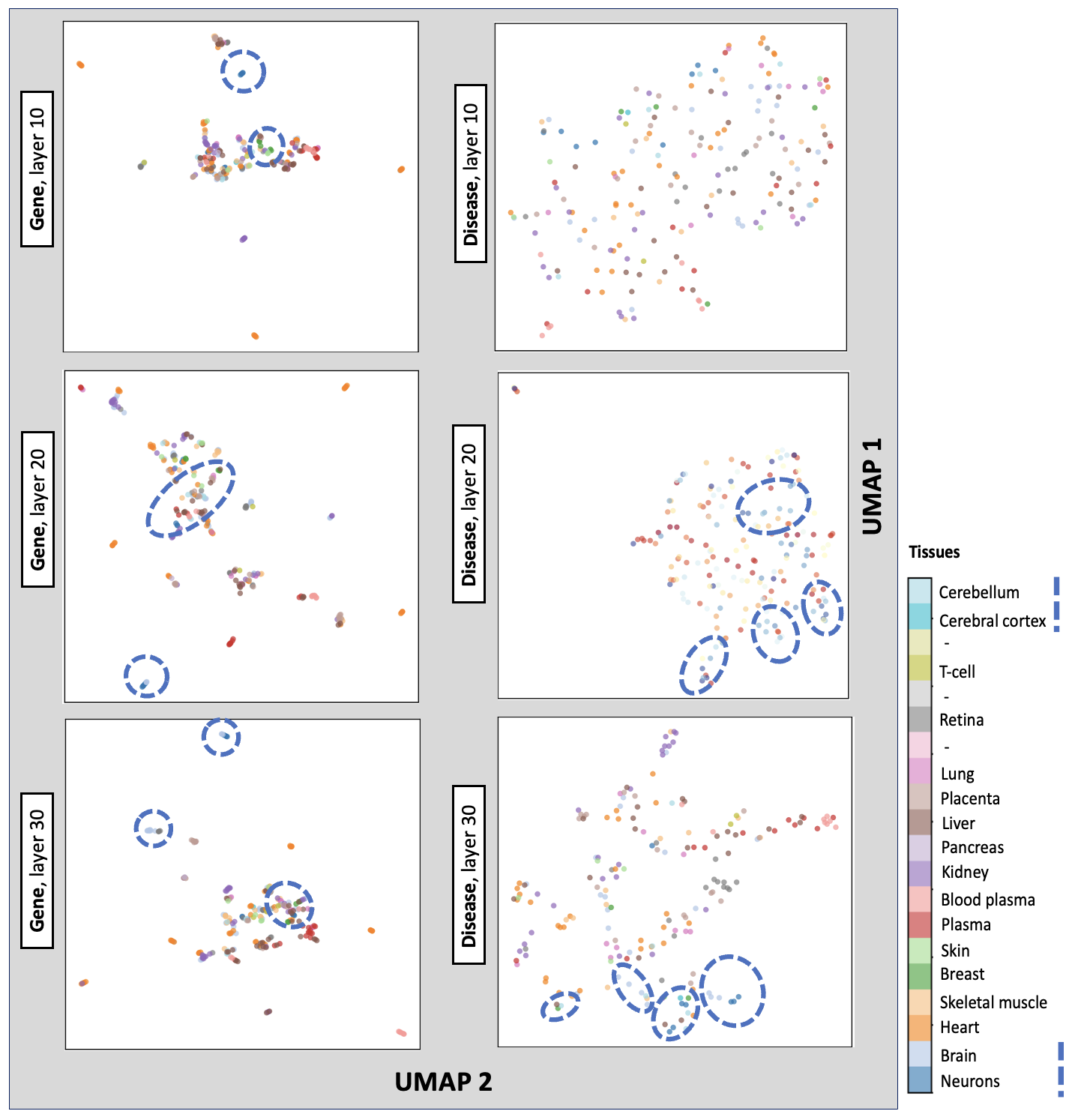}}
  \captionof{figure}{Latent text representations of gene entities and phenotype entities in GP-GPT(base). The subplots show the embedding implant in the 2D UMAP space, for layer 10th, layer 20th, and layer 30th. Blue circles indicate the identified clusters of neuron-related tissues. }
  \label{fig9}
\end{minipage}

\begin{minipage}[b]{1.03\linewidth}
  \centering
  \centerline{\includegraphics[width=\textwidth,height=17.8cm]{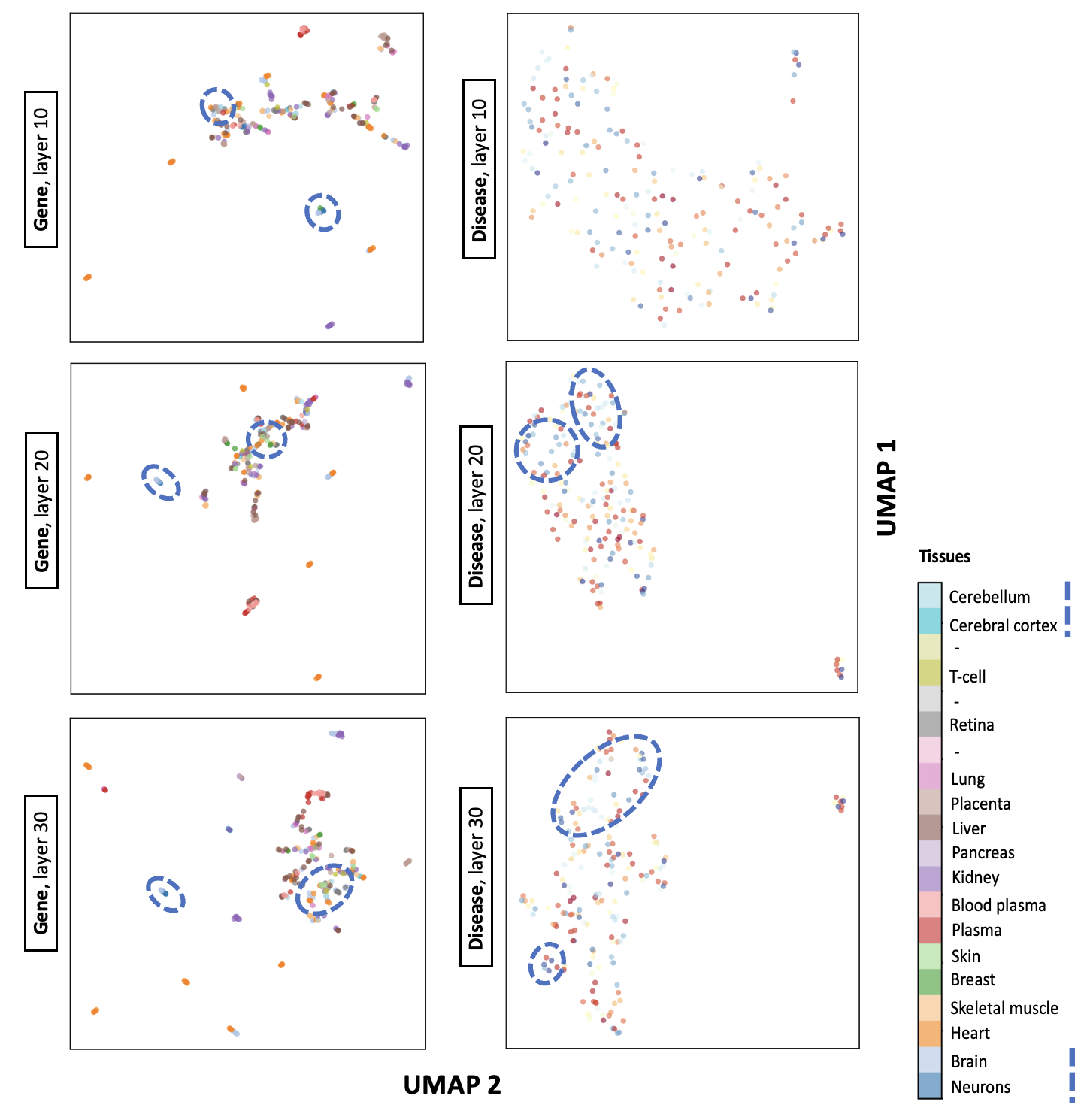}}
  \captionof{figure}{Comparable zero-shot embedding of text representations of gene entities and phenotype entities in original Llama3.1(8B). The subplots show the embedding implant in the 2D UMAP space, for layer 10th, layer 20th, and layer 30th. Blue circles indicate the identified clusters of neuron-related tissues. }
  \label{fig10}
\end{minipage}

\section{Discussions and Conclusion}
\subsection{Limitations}
Our study faces a few limitations. Considering the current research trends and inherent issues in the realms of genetics, genomics, transcriptome, proteome, metabolome, and the general biomedical domain, the primary challenges of large language models lie in the following key areas.

\subsubsection{Dataset Expansion}
The data used in this study covers multiple data sources, including dbGaP, UniPort, DisGeNet, NCBI, and OMIM. Among these datasets, the employed bio-text data are mainly based on OMIM. Although the training corpus is composed of 4401 pairs of phenotype and gene, the construction of training contextual data is heavily dependent on high-quality textual data, which is lacking for most genes and diseases. With this limitation of data sources, there is demand of generating more reliable genomics textual data from other structure dataset. Inspired by the network-based disease gene prediction research, using the network data of protein-protein interaction (PPI), gene expression, gene ontology, and other domain knowledge to generate large relational-genomics text data is an available way. Based on conventional relationships often observed in bio-networks, e.g. dominating, inhibiting, and promoting, the relation network of bio-factors can provide rich text data following a well-designed graph-to-text interpreting system. To be noted that, such systems are usually based on LLMs as well~\cite{chai2023graphllm, swamy2021interpreting, hu2023beyond}.

\subsubsection{Tokenization of Genomics Entities}
Tokenization is one of the essential steps in the success of the training language model. To simplify the model training, we adopt the character-based tokenization method from the original Llama2, and Llama3 for all the genomics entities. One reason for this is that the model is only focusing on the name-entities features and relation description, which both are based on natural languages. Another reason is that Llama2 and Llama3 employ a character-based method for the numerical tokenization approach, which has been shown to be more effective across various arithmetic tasks. In practice, these tokenizers work well for the current stage of GP-GPT. However, a from-scratch pre-training can be necessary for a better representation of genetics, genomics, and other levels of bio-factors, especially when more features from the sequence are involved. Previous works~\cite{bryant2022improved, abramson2024accurate, pei2023biot5} have provided available instances for bioinformatics costumed tokenizor. Due to the computation cost, we decide to put such endeavours into our further improvement of GP-GPT.

\subsubsection{Model Performance and Evaluation}
In terms of question-answer text synthesis, the model occasionally exhibits semantic repetition at the document level, tends to lose coherence over the questions, and sometimes includes non-sequitur sentences or paragraphs. Our experiments only include the instruction fine-tuning following the convention of the generative language model and do not include self-regression architectures or other denoising training objectives. Lack of direct training on fill-in-the-blank tasks or considering bi-directionality self-regression leads to potentially worse performance on parsing a long passage and then generating a very short answer. In practice, that could be observed at relatively low precision metrics for all LLMs tested in this study, compared with other knowledge graph-based models.

Furthermore, the evaluation methods for such genomics language models are still underdeveloped. The direct issue to this end is that the missing of standard benchmarks in this area. On one hand, traditional metrics such as BLEU~\cite{papineni2002bleu}, and ROUGE~\cite{lin2004rouge} do not fit the defined research aim in this study, thus the metrics lack the necessary focus on certain bio-factor entities. In this study, we use modified BLEU which adjusts the weighting for genomics keywords to mediate this issue. Both the original BLEU score and modified BLEU score are reported for GP-GPT, leading to a more comprehensive estate of model performance. On the other hand, a stand-alone test set for evaluation becomes a problem for this study. That is due to the aim of the model being to build a fundamental language AI system to interpret real-world genetic medical relations. Our solution in the study is based on internal validation on a random select genomics relationship which has been verified by biological experiments.

\subsection{Future Perspectives}


\subsubsection{Multi-Modality}
Our proposed large language models are not grounded in incorporating other domains of data, such as biological sequence data and medical images. The wide applicability of transformer architecture~\cite{vaswani2017attention} facilitates modality fusion. Recent research in computer vision and NLP also demonstrates significant benefits in combining multiple modalities to achieve application success~\cite{radford2021learning,liu2024sora,team2024chameleon,lee2023multimodality,dou2023towards,xiao2023instruction}. Domain-specific models are natural platforms for incorporating data from other modality to improve and scale practical usability~\cite{liu2024understanding,liu2024surviving,li2023artificial,wang2024large,zhao2022embedding,liu2023radonc,cai2023exploring,cai2023multimodal,liu2023tailoring,tan2023promises}.

Bio-sequence data counts for the most dominant part of all available data in bioinformatics studies. One of the efficient ways to combine the bio-sequence data and LLMs is utilizing the parallel architecture to place both the sequence-specific model and language-specific model~\cite{xu2023protst}. Using specific tokenizers to encode particular sequence features is another practical method~\cite{heinzinger2023bilingual}. In future versions of GP-GPT, the sequence data from DNA sequences, gene variances, and gene products(small-RNA, protein) will be included in the framework.

Imaging genetics offers the possibility of building connections between genetics insights into the biology of complex diseases and functions~\cite{bigos2010imaging,bi2023community,bi2023structure}. LLM systems can largely benefit the representation of gene variants and candidate phenotypes, thus providing a qualitative description of domain knowledge. By utilizing the image encoder, bio-factor entity encoder, and variants encoder, this framework can capture the intricate interplay in semantic embedding space, which leads to more comprehensive association identifications.


\subsubsection{Dementia-related Genotype-Phenotype Associations Identification} 
Dementia is a major public health challenge due to its profound impact on individuals, families, and healthcare systems globally. Dementia's etiology is complex, involving a combination of genetic and environmental factors~\cite{kwok2020complex}. Alzheimer's Disease (AD)~\cite{montine2012national,zhang2024disease2vec,zhang2021deep}, the most common form of dementia, is characterized by the accumulation of amyloid plaques and neurofibrillary tangles in the brain. Over the past few decades, genetics has become increasingly pivotal in AD study~\cite{bateman2011autosomal,saykin2015genetic}, starting with the discovery of rare mutations in the APP (amyloid precursor protein), PSEN1 (presenilin-1), and PSEN2 (presenilin-2) genes, which are linked to early-onset autosomal dominant forms of the disease. The association of the APOE (apolipoprotein E) $\epsilon$-4 allele with sporadic or late-onset AD has further highlighted the genetic underpinnings of the disease. Phenotypic heterogeneity, or variability in disease presentation, is also evident in AD ~\cite{kwok2020complex}. For instance, certain PSEN1 mutations are associated with distinct clinical features such as spastic paraparesis and degeneration of the corticospinal tract~\cite{menendez2004pathological,shepherd2004positional}. Similarly, APP mutations can lead to lobar haemorrhages associated with cerebral A$\beta$40 angiopathy, which showcases the diverse manifestations of AD ~\cite{grabowski2001novel}. Dementia with Lewy bodies (DLB) is another form of progressive dementia marked by the presence of Lewy bodies—abnormal protein deposits in the brain ~\cite{lin2019diffuse}. These deposits disrupt brain chemistry, leading to memory, cognitive, movement, behavioural, and mood disturbances. DLB shares clinical features with both AD and Parkinson's disease, complicating its diagnosis. Genetic research has identified several mutations associated with Lewy body diseases, including mutations in LRRK2, SNCA, PLA2G6, and C19ORF12, as well as GBA variants ~\cite{schneider2017neuropathology}, which significantly increase the risk of developing these conditions. Genes and their variants are not isolated entities; instead, they function within broader pathways and networks, leading to a range of downstream effects across multiple biological levels, including the transcriptome, proteome, and metabolome. For instance, previous studies have shown that the pro-apoptotic gene FASTKD2 is associated with better episodic memory performance. This association relates to a memory-associated SNP linked to lower FASTKD2 mRNA expression (transcript), reduced cerebrospinal fluid levels of fas-mediated apoptotic factors (protein), and greater hippocampal volume and gray matter density (brain anatomy)~\cite{ramanan2015fastkd2}. Therefore, studying genotype-phenotype associations is essential for advancing research, improving diagnosis, and developing targeted therapies that consider the individual variability in patients.
GP-GPT shows significant promise in addressing this issue. It excels at genotype-phenotype mapping, enabling the analysis of genetic impacts on various phenotypes. By integrating genotype and phenotype into a unified embedding space, GP-GPT positions genes with similar pathological roles close to one another. This approach offers a novel method for identifying new candidate high-risk genes by exploring the continuous embedding space around known pathogenic genes.

\subsection{Conclusion}
In this study, we introduce GP-GPT, a specialized large language model framework for genetic-phenotype knowledge representation and analysis, which incorporates multiple gene, protein, and biomedical data sources. The training dataset integrates genomics and genetic medical data from key databases such as OMIM, DisGeNET, UniProt, and dbGaP to compile a consortium of both structured and unstructured data, and organizes this raw data into textual corpus. Based on the Llama model family, we developed genomic-specific versions of GP-GPT in three sizes: GP-GPT Small (based on Llama 2 7B), GP-GPT Base (based on Llama 3 8B), and GP-GPT Large (based on Llama3 70B). We evaluated the model on multiple tasks, including genetic medical question answering and genomic relationship determination. In such tasks, GP-GPT outperformed current state-of-the-art large language models, including Llama2/3 and GPT-4. Additionally, a comparison of the GP-GPT variants (Small, Base, Large) revealed a significant impact of model size on fine-tuning outcomes.

Overall, the fine-tuned models with different configurations(GP-GPT base, GP-GPT small, and GP-GPT large) all delivered superior performance on genomics QA tasks, providing more reliable answers to genetic medical questions and more accurate entity identifications. Furthermore, we demonstrated that fine-tuning large language models on genomics-specific training data leads to improved hidden representations of genetic medical entities. This indicates that the model has gained a deeper understanding of the complex relationships among bio-factor entities. For instance, compared to the original pre-trained model, when focusing on the neuron-brain-related genes and diseases, clear embedding clusters emerged for GP-GPT models' entity embeddings. 
GP-GPT consistently offers an effective framework for studying the representations of genetic medical concepts and genomic bio-factors. 

As the first large language model to describe the relationships among multi-level bio-factors, GP-GPT models show potential for further refinement and the capacity to become a foundational model in genomics and medical genetics. One key application is AI-assisted genetic disease prediction. By integrating extensive knowledge, the model can serve as an AI assistant in diagnosing gene-related diseases in individual clinical cases. Additionally, in large-scale studies of associations between diseases and genetic variants, the reasoning behind results often depends on carefully designed experiments and rigorous statistical corrections in large cohorts. GP-GPT can enhance this process by offering strong probabilistic priors, particularly in cases of small sample sizes and extreme data imbalances.

\section{Model and Data Availability}

The model in this work is openly available at  
\url{https://huggingface.co/IanL10/GP-GPT}.  

Processed datasets for training is openly available at:  
\url{https://huggingface.co/datasets/IanL10/Genomic-text}.

\bibliography{LLM_GdP}
\bibliographystyle{unsrt}

\end{document}